\documentclass[conference]{IEEEtran}
\IEEEoverridecommandlockouts
\usepackage{cite}
\usepackage{amsmath,amssymb,amsfonts}
\usepackage{graphicx}
\usepackage{listings}
\usepackage{booktabs} 
\usepackage{longtable}
\usepackage{supertabular}
\usepackage{easyReview}
\usepackage{textcomp}
\usepackage{xcolor}
\usepackage{pythonhighlight}
\usepackage{multicol}
\usepackage{tabularray}  
\usepackage{amssymb}
\usepackage{lipsum}
\usepackage[ruled,vlined]{algorithm2e}
\usepackage[noend]{algpseudocode}
\usepackage{amsmath}
\usepackage{multirow}
\usepackage{easyReview}
\usepackage{soul}
\usepackage{arydshln}
\usepackage{graphicx}
\usepackage{listings}
\usepackage{hyperref}
\usepackage{url}
\usepackage{fontawesome} 
\newcommand{\huggingface}{\includegraphics[height=2ex]{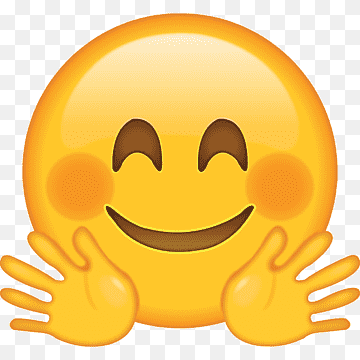}}
\makeatletter
\newcommand{\linebreakand}{%
  \end{@IEEEauthorhalign}
  \hfill\mbox{}\par
  \mbox{}\hfill\begin{@IEEEauthorhalign}
}
\makeatother
\def\BibTeX{{\rm B\kern-.05em{\sc i\kern-.025em b}\kern-.08em
T\kern-.1667em\lower.7ex\hbox{E}\kern-.125emX}}
\usepackage{todonotes}
\usepackage{draftwatermark}
\SetWatermarkText{Accepted in IEEE BigData 2025}
\SetWatermarkScale{0.3}        
\SetWatermarkColor[gray]{0.8} 

\begin{document}
\title{
multiMentalRoBERTa: A Fine-tuned Multiclass Classifier for Mental Health Disorder

}
\IEEEoverridecommandlockouts
\IEEEpubid{\begin{minipage}[t]{\textwidth}\ \\[10pt]
\end{minipage}}

\author{\IEEEauthorblockN{K M Sajjadul Islam}
\IEEEauthorblockA{{Marquette University, WI, USA} \\
sajjad.islam@marquette.edu}
\and
\IEEEauthorblockN{John Fields}
\IEEEauthorblockA{{Marquette University, WI, USA} \\
john.fields@marquette.edu}
\and
\IEEEauthorblockN{Praveen Madiraju}
\IEEEauthorblockA{{Marquette University, WI, USA} \\
praveen.madiraju@marquette.edu}
}
\maketitle 

\begin{abstract}
The early detection of mental health disorders from social media text is critical for enabling timely support, risk assessment, and referral to appropriate resources. This work introduces multiMentalRoBERTa, a fine-tuned RoBERTa model designed for multiclass classification of common mental health conditions, including stress, anxiety, depression, post-traumatic stress disorder (PTSD), suicidal ideation, and neutral discourse. Drawing on multiple curated datasets, data exploration is conducted to analyze class overlaps, revealing strong correlations between depression and suicidal ideation as well as anxiety and PTSD, while stress emerges as a broad, overlapping category. Comparative experiments with traditional machine learning methods, domain-specific transformers, and prompting-based large language models demonstrate that multiMentalRoBERTa achieves superior performance, with macro F1-scores of 0.839 in the six-class setup and 0.870 in the five-class setup (excluding stress), outperforming both fine-tuned MentalBERT and baseline classifiers. Beyond predictive accuracy, explainability methods, including Layer Integrated Gradients and KeyBERT, are applied to identify lexical cues that drive classification, with a particular focus on distinguishing depression from suicidal ideation. The findings emphasize the effectiveness of fine-tuned transformers for reliable and interpretable detection in sensitive contexts, while also underscoring the importance of fairness, bias mitigation, and human-in-the-loop safety protocols. Overall, multiMentalRoBERTa is presented as a lightweight, robust, and deployable solution for enhancing support in mental health platforms.
\end{abstract}
\vspace{1ex}

\begin{IEEEkeywords}
Mental Health, Explainable AI (XAI), Classification, Fine-tune, Prompt Engineering, BERT, SBERT, RoBERTa, LLM, GPT, DeepSeek 
\end{IEEEkeywords}

\section*{}
\vspace{-3ex}
{\faGithub \hspace{0.1ex} Code\footnote{https://github.com/sajjadIslam2619/mental-health-disorder-analysis}}
\hspace{1ex} {\huggingface \hspace{0.1ex} Models\footnote{https://huggingface.co/collections/SajjadIslam/multimentalroberta-models}}

\section{Introduction}
The World Health Organization defines mental health as ``a state of mental well-being that enables people to cope with the stresses of life, realize their abilities, learn and work well, and contribute to their community," and of mental disorders as clinically significant disturbances in cognition, emotional regulation, or behavior that cause distress or functional impairment\cite{Mental-health}. World Health Organization Mental disorders are common: In 2019, approximately one in eight people worldwide (970 million) lived with a mental disorder, with anxiety and depressive disorders the most prevalent; early pandemic estimates showed sharp increases in both conditions \cite{Mental-disorders}. The public-health burden is underscored by suicide, which claims more than 720,000 lives annually and ranks as a leading cause of death among young people (ages 15 - 29) globally\cite{Suicide}. 

In modern times, many individuals turn to social media platforms and peer-support forums to share experiences and seek help for mental health concerns. Among the most commonly discussed conditions are depression (characterized by persistent sadness, hopelessness, and loss of interest in daily activities \cite{Mental-disorders}), anxiety disorders (involving excessive fear or worry that interferes with daily functioning \cite{Mental-disorders}) and post-traumatic stress disorder (PTSD) (marked by intrusive memories, avoidance, and hyperarousal following traumatic events \cite{PTSD}). Another frequent concern is suicidal ideation, defined as persistent thoughts about or planning of taking one’s own life \cite{Suicide}. Stress is also widely examined in online contexts, though it is not formally classified as a mental disorder. Instead, it is a psychological state that can contribute to or worsen mental illnesses, with the National Center for Complementary and Integrative Health (NCCIH) noting that long-term stress may increase vulnerability to conditions such as depression and anxiety \cite{Stress}. In our dataset, we also observe that stress occupies an intermediate space between mental health disorders and neutral data (see Figure~\ref{Fig:class-cluster}).

Classifying mental health disorders in peer-support systems and online forums is critical, as these platforms are often where individuals first disclose symptoms due to accessibility and anonymity. Accurate classification enables timely support, risk detection, and referral to appropriate resources, which is particularly important for conditions such as depression and suicidal ideation. Advances in artificial intelligence (AI) and machine learning (ML) provide effective tools for this task, as they can analyze large volumes of social media data, identify linguistic and behavioral patterns, and improve early detection of mental health problems \cite{abd2020application}. Beyond scalability, AI has shown promise in enhancing diagnostic accuracy and addressing global shortages of mental health professionals, offering a cost-effective way to expand access to care \cite{minerva2023ai}. A key difficulty arises from the contrast between binary and multi-class classification: in binary setups, random guessing yields a 50\% accuracy, whereas in a six-class problem, it drops to about 16.7\% (1/6). This gap underscores the heightened complexity of multi-class classification, where models face a more demanding decision space with a narrower margin for error \cite{fields2024integrating}. Building on this context, the present work contributes to advancing multiclass detection of mental health disorders through the following aims:

Aim 1: Curate and explore datasets of mental health discourse, including quantitative analysis of class similarity and overlap across conditions. (see sections: \ref{sec-data-collection}, \ref{sec-data-processing} )

Aim 2: Develop and evaluate models for mental disorder detection, comparing baseline methods, domain-specific transformers, and large language models. (see sections: \ref{sec-Traditional-ML}, \ref{sec-transformers}, \ref{sec-prompt}, \ref{result-analysis})

Aim 3: Perform explainability and safety analyses, focusing on depression and suicidal ideation to identify key signals and assess reliability. (see sections: \ref{explainability-analysis}, \ref{sec-bias})

\section{Related Work}
\subsection{Peer Support and Social Media in Mental Health Research}\label{sec-peer-support}
Online peer-support platforms and social media communities have become important spaces for individuals to share experiences, seek advice, and disclose symptoms related to mental health. Platforms such as Reddit, Twitter, and Facebook not only facilitate supportive exchanges but also provide rich data sources for analyzing depression, stress, anxiety, PTSD, and suicidality \cite{garg2023mental}. These communities help reduce stigma, encourage help-seeking, and often provide early signals of distress, complementing traditional clinical pathways. Self-disclosure plays a central role, as individuals candidly share vulnerable experiences. Research shows that self-disclosure supports psychological well-being and therapeutic progress \cite{balani2015detecting}, and studies of Reddit forums indicate that highly self-disclosing posts, though often receiving fewer upvotes, attract faster and more extensive responses, reflecting strong community support for vulnerable users.

\subsection{Computational Approaches for Mental Health Disorder Detection}\label{sec-computational-approach}
Early computational approaches to mental health detection from non-clinical texts relied on traditional machine learning methods such as Support Vector Machines (SVM), logistic regression, and decision trees, using lexical features (e.g., bag-of-words, TF-IDF), psycholinguistic lexicons like LIWC, and basic behavioral indicators. These methods showed feasibility but struggled with generalization and capturing nuanced semantics \cite{calvo2017natural}. With deep learning, research shifted to CNNs, RNNs, and attention-based models. Ji et al. \cite{ji2022suicidal} introduced an Attentive Relation Network combining sentiment lexicons, latent topics, and embeddings to model relationships between suicidal ideation and related disorders, outperforming strong baselines across datasets. Subsequent work advanced the field with deep contextualized representations. Jiang et al. \cite{jiang2020detection} introduced a large Reddit-based dataset spanning eight psychiatric disorders and showed that BERT-based models substantially outperformed LIWC baselines, with user-level modeling proving more effective than post-level classification. Still, multiclass classification remained difficult due to overlap and comorbidity. More recent studies explored multimodal and knowledge-augmented models: Toto et al. \cite{toto2021audibert} proposed AudiBERT, integrating audio and text with dual self-attention for depression screening, while Yang et al. \cite{yang2022mental} developed KC-Net, which combines commonsense mental state knowledge with supervised contrastive learning for stress and depression detection.

\subsection{Explainability in Mental Health Models}\label{sec-xai}
The application of AI to mental health detection requires not only high predictive accuracy but also explainability, as clinicians, peer support specialists, and users must trust model outputs in high-stakes contexts. Recent surveys distinguish between explainable AI (XAI), which provides post-hoc explanations, and interpretable AI (IAI), where models are inherently transparent \cite{huang2024explainable}. Most healthcare NLP work emphasizes local post-hoc methods such as LIME, SHAP, and attention visualization, while global interpretability remains limited. In mental health, explainability is increasingly linked to clinical grounding: Bao et al. \cite{bao2024explainable} showed that transformer-based models can detect depressive symptoms and generate explanations aligned with clinical assessments (e.g., PHQ-9), with extractive explanations proving more reliable than abstractive ones. Joyce et al. \cite{joyce2023explainable} similarly stressed the need for understandability - combining transparency and interpretability - through their TIFU framework, which aligns explanations with psychiatric reasoning.

\subsection{Advancing Mental Health Classification}\label{sec-present-study}

Recent work such as MentaLLaMA\cite{yang2023mentalllama} has advanced the field by introducing an instruction-tuned large language model for interpretable mental health analysis. Its major focus is on text generation, producing both classifications and explanatory rationales in a single step. To achieve this, MentaLLaMA requires fine-tuning LLaMA2 models with 7B or 13B parameters on the IMHI dataset. While it demonstrates strong generalization and generates human-readable explanations, its evaluation also showed that discriminative models like MentalBERT\cite{ji2022mentalbert} still achieve superior classification accuracy across several benchmarks. MentalBERT addressed the gap of lacking domain-specific pretrained language models by continuing pretraining on large-scale Reddit corpora from mental health communities. The resulting models improved performance over general-purpose BERT and RoBERTa and provided the first publicly available pretrained embeddings for the domain. To be effectively applied, however, these embeddings still require fine-tuning on specific downstream classification tasks. Most evaluations focused on binary detection tasks such as depression vs. control or suicidal vs. non-suicidal, with only limited exploration of multiclass classification. In addition, the work did not explicitly address explainability or the correlation of overlapping disorders.

The present study extends this line of research by introducing multiMentalRoBERTa, a fine-tuned transformer model tailored for multiclass classification of common mental health disorders, including stress, anxiety, depression, PTSD, and suicidality. Unlike resource-intensive instruction-tuned LLMs or pretrained models requiring additional fine-tuning, this model is already fine-tuned, lightweight, and readily deployable. In addition, it integrates data exploration, correlation analysis, and explainability methods, including token-level attributions and key phrase extraction, to improve transparency and trust. This positions multiMentalRoBERTa as a practical and effective approach for integration into peer-support platforms, social media monitoring, and online mental health disorder detection systems.

\section{Methodology}
\subsection{Data Collection \& Curation}\label{sec-data-collection}

\textbf{Reddit Depression Dataset}\cite{pirina2018identifying} - is constructed from multiple balanced subsets of Reddit and forum posts, with 400 documents per class. Labels are not human-annotated but are assigned automatically based on subreddit membership. Posts from the Depression Support subreddit (DS) are treated as depression cases, while additional subsets such as posts from users with self-declared depression in other subreddits (DO) are also included. Control subsets from subreddits like Breast Cancer (BC), Family/Friendship (FF), and non-depression forums (DND) are provided in the original dataset but are not used in this research. For the present study, only the DS and DO subsets are selected, as they provide self-disclosure and broader depression-related content most relevant to the research objective.

\textbf{Dreaddit Dataset}\cite{turcan2019dreaddit} - is constructed from Reddit posts collected between January 2017 and November 2018 across multiple subreddits. It is human-annotated through Amazon Mechanical Turk, where each sampled segment is labeled for the presence of stress. For this research, only posts related to stress, anxiety, and PTSD are collected. Stress data are obtained from subreddits focused on abuse, financial need, and social relationships, while anxiety posts are taken from r/anxiety, and PTSD posts are taken from r/ptsd. The dataset therefore provides human-labeled user-generated text that captures mental health challenges across these three categories, which are retained for analysis.

\textbf{The Stress Annotated Dataset (SAD)} \cite{mauriello2021sad} - is composed of 6,850 SMS-like sentences that are generated through chatbot interactions, crowdsourcing, and targeted web scraping from an online repository. Each sentence is labeled by multiple annotators with one or two stressor categories and a severity rating, and nine final categories are established, including work, school, financial problems, family issues, health or fatigue, social relationships, emotional turmoil, everyday decision making, and other. For this research, only the stress-labeled sentences are retained to ensure that the dataset is focused exclusively on texts that express stressful experiences.

\textbf{Reddit Suicide and Depression (SDCNL) Dataset} \cite{haque2021deep} - is constructed by collecting 1,895 posts from two mental health subreddits, where posts from r/SuicideWatch are labeled as suicidal and posts from r/Depression are labeled as depressed. Each post is retained with its original text and subreddit-based label, although the annotations are inherently noisy due to reliance on self-reporting. For this research, only the suicidal and depression data are selected to enable a focused analysis on distinguishing between depressive expression and suicidal ideation.

\textbf{Neutral Data Collection} - to provide a neutral comparison class, posts without any mental health context are collected from the subreddits `ExplainLikeImFive' and `TodayILearned'. This “None” data is important as it represents everyday discussions unrelated to psychological conditions, allowing the models to better distinguish between clinically relevant and ordinary online discourse. The dataset is curated by filtering posts with a maximum of 400 words and removing duplicates to ensure quality and consistency. This non-mental health category serves as a necessary baseline for evaluating classification performance against Stress, Anxiety, Depression, PTSD and Suicidal content.

\begin{table}[tb]
\caption{Class Distribution and Word Count Statistics}
\label{table-label-wordcount}
\centering
\setlength{\tabcolsep}{3pt}
\begin{tabular}{lcccc}
\toprule
Class & Count & Min Words & Max Words & Median Words \\
\midrule
Stress      & 2274 & 10  & 310 & 26  \\
Anxiety     & 416  & 1   & 260 & 84  \\
Depression  & 2322 & 10   & 400 & 119 \\
PTSD        & 414  & 1   & 256 & 80  \\
Suicidal    & 838  & 1   & 398 & 86  \\
None        & 706  & 1   & 359 & 52  \\
\bottomrule
\end{tabular}
\end{table}

Overall, stress and depression emerge as the dominant classes in the collected corpus. To maintain balance and ensure data quality, posts containing fewer than 10 words or more than 400 words are trimmed from the dataset. The final distribution of classes along with their word count statistics is summarized in Table~\ref{table-label-wordcount}.

The use of social media data in mental health research also raises ethical challenges. While posts are technically public, users often perceive them as semi-private, with limited awareness of data permanence and computational analysis \cite{benton2017ethical}. Ethical guidelines emphasize informed consent where possible, de-identification, secure storage, and sensitive handling of disclosures. Scholars recommend practices such as paraphrasing examples and avoiding cross-platform linkages that could re-identify individuals. Importantly, for this work no user-identifiable information is utilized, and the classification models developed are not capable of exposing sensitive information, thereby aligning with these ethical principles.

\subsection{Data Exploration}\label{sec-data-processing}

Data exploration is carried out to understand the structure of the corpus and the separability of different mental health categories prior to model development. This step is important as it provides insights into data quality, embedding suitability, and potential overlaps between classes, all of which influence the robustness of subsequent modeling.

Clustering is performed using different embeddings, and the quality of representations is evaluated through three metrics: Adjusted Rand Index (ARI), which measures the agreement between clustering results and ground-truth labels; Normalized Mutual Information (NMI), which quantifies how much information is shared between clusters and true labels; and Silhouette score, which assesses cohesion within clusters and separation between clusters. As shown in Table~\ref{table-embedding-performance}, RoBERTa provides the best embedding quality across all three metrics. Consequently, this embedding is used for subsequent analysis. Using RoBERTa representations, class distribution is examined, and cosine similarity between centroids is computed.

\begin{table}[tb]
\caption{Embedding-based Clustering Evaluation}
\label{table-embedding-performance}
\centering
\begin{tabular}{lccc}
\toprule
Model & ARI & NMI & Silhouette \\
\midrule
BERT & 0.109 & 0.176 & 0.0043 \\
SBERT      & 0.251 & 0.336 & 0.0246 \\
RoBERTa  & 0.255 & 0.347 & 0.0327 \\
\bottomrule
\multicolumn{4}{l}{*see section \ref{sec-transformers} for embedding models}
\end{tabular}
\end{table}

\begin{figure*}[tb]
\centerline{\includegraphics[width=0.91\textwidth]{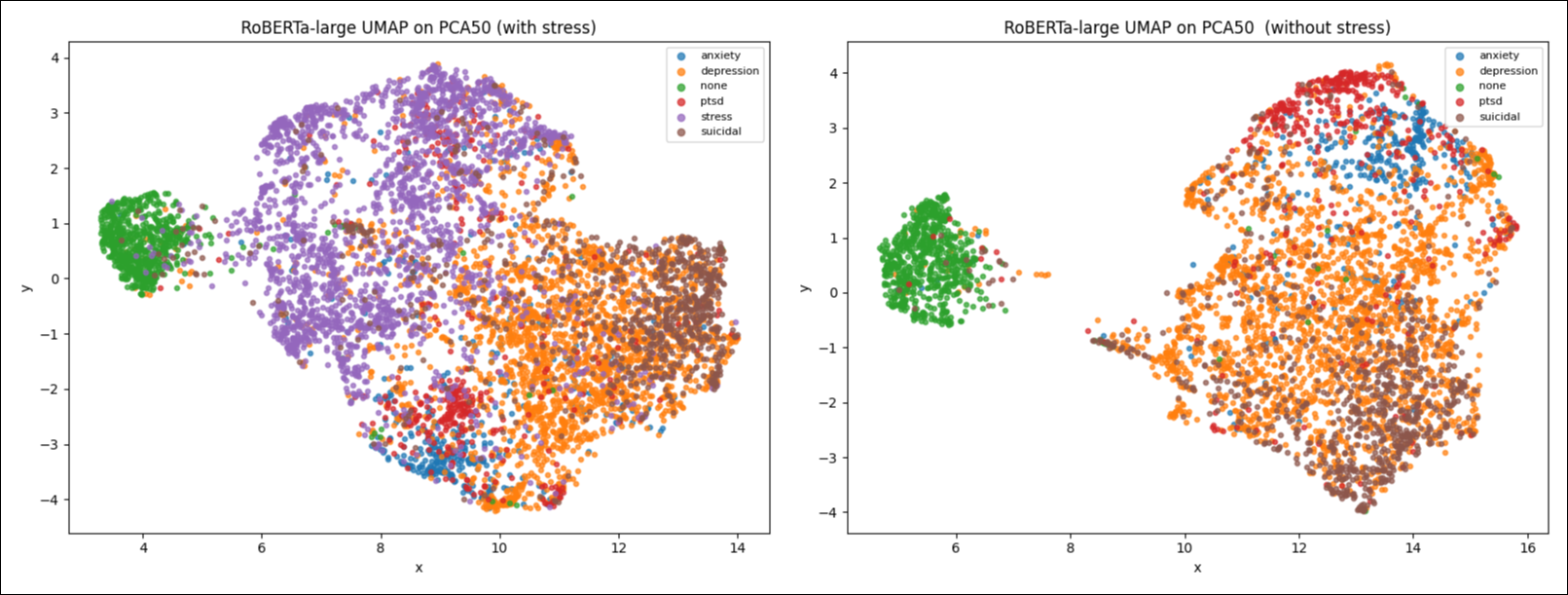}}
\caption{Mental Health Disorder Data Cluster in Embedding Space (RoBERTa Embedding)}
\label{Fig:class-cluster}
\end{figure*}

\begin{table}[tb]
\caption{Cluster Distribution by Classes (RoBERTa Embedding)}
\label{table-cluster-distribution}
\centering
\begin{tabular}{lrrrrrr}
\toprule
\multirow{3}{*}{Class} & \multicolumn{6}{c}{\textbf{Cluster}} \\
\cmidrule(lr){2-7}
  & 0 & 1 & 2 & 3 & 4 & 5 \\
\midrule
Stress      & 98  & 375 & 47   & 896 & 752 & 106 \\
Anxiety     & 17  & 9   & 30   & 36  & 50  & 274 \\
Depression  & 68  & 36  & 1185 & 202 & 174 & 657 \\
PTSD        & 7   & 4   & 31   & 13  & 159 & 200 \\
Suicidal    & 50  & 14  & 590  & 85  & 48  & 51 \\
None        & 694 & 0   & 0    & 2   & 1   & 10 \\
\bottomrule
\end{tabular}
\end{table}

Figure~\ref{Fig:class-cluster} provides a visual representation of how different mental health conditions are positioned in the embedding space, and the patterns observed are consistent with the clustering and correlation results. Table~\ref{table-cluster-distribution} shows how true labels are distributed across clusters when RoBERTa embeddings are applied, while Table~\ref{table-correlation-matrix} quantifies correlations among the classes. The visualization and tables together indicate that None is concentrated almost entirely in a single cluster, suggesting strong separability, whereas Stress and Depression are spread across multiple clusters. Notably, Suicidal and PTSD overlap with Depression, reflecting linguistic similarity. The correlation matrix confirms that the highest correlation (0.929) occurs between Depression and Suicidal, followed by strong associations between Anxiety and PTSD (0.843) and between Stress and PTSD (0.748). In contrast, the None class shows very low correlation with all other categories, reinforcing its distinct nature. The widespread distribution of Stress across clusters can be explained by its role as a broad underlying factor that contributes to multiple psychiatric conditions. Prior work \cite{davis2017neurobiology} highlights that chronic stress is strongly associated with the onset of major depressive disorder, PTSD, and other serious mental illnesses, which supports the observation that stress does not remain isolated but overlaps with other mental health categories. These overlaps may be attributed to shared vocabulary, co-occurring symptoms, and thematic similarities across conditions, which can blur class boundaries. 

Finally, based on the outcomes of this exploration phase, the dataset is prepared for model training under two experimental setups: one including Stress as a separate class (six-class classification) and one excluding Stress (five-class classification). In the following sections on Result Analysis (\ref{result-analysis}) and Explainability  Analysis (\ref{explainability-analysis}), these relationships are examined in greater depth to better understand model performance and interpretability.

\begin{table}[tb]
\caption{Correlation Matrix among Mental Health Conditions (RoBERTa Embedding)}
\label{table-correlation-matrix}
\centering
\setlength{\tabcolsep}{3pt}
\begin{tabular}{lcccccc}
\toprule
 & Stress & Anxiety & Depression & PTSD & Suicidal & None \\
\midrule
Stress     & 1.000 & 0.680 & 0.714 & 0.748 & 0.698 & 0.030 \\
Anxiety    & 0.680 & 1.000 & 0.775 & 0.843 & 0.666 & 0.122 \\
Depression & 0.714 & 0.775 & 1.000 & 0.730 & 0.929 & 0.163 \\
PTSD       & 0.748 & 0.843 & 0.730 & 1.000 & 0.678 & 0.058 \\
Suicidal   & 0.698 & 0.666 & 0.929 & 0.678 & 1.000 & 0.114 \\
None       & 0.030 & 0.122 & 0.163 & 0.058 & 0.114 & 1.000 \\
\bottomrule
\end{tabular}
\end{table}

\subsection{Traditional Machine Learning Models}\label{sec-Traditional-ML}
Logistic Regression and Linear SVM are employed as baseline models due to their simplicity, efficiency, and well-established effectiveness in text classification tasks. Instead of relying on traditional word embeddings such as TF-IDF, these models are trained on dense representations generated by the RoBERTa model. This setup ensures that even the baseline classifiers benefit from contextual embeddings, while still serving as a reference point against which more advanced transformer-based and prompting approaches can be evaluated.

\subsection{Transformer Fine-Tuned Models}\label{sec-transformers}

\textbf{BERT (bert-large-uncased)}\footnote{https://huggingface.co/google-bert/bert-base-uncased} - is a transformer-based language model pretrained on large-scale text corpora using masked language modeling and next sentence prediction. The bert-large-uncased variant consists of 24 layers with 340 million parameters, enabling strong contextual understanding of text. In this study, the model is fine-tuned for multi-class classification of mental health conditions, adapting its pretrained representations to the task-specific label space.

\textbf{SBERT (all-mpnet-base-v2)}\footnote{https://huggingface.co/sentence-transformers/all-mpnet-base-v2} - is a variant of transformer models designed to generate high-quality sentence embeddings. The all-mpnet-base-v2 model is based on MPNet-base with 12 layers, 768 hidden dimensions, 12 attention heads, and approximately 109 million parameters. It produces 768-dimensional normalized embeddings that are particularly effective for semantic similarity, clustering, and retrieval tasks. While primarily optimized for embedding-based applications, for this work it is also fine-tuned here for classification, benefiting from its strong performance in representing sentence-level semantics and capturing nuanced relationships between mental health categories.

\textbf{MentalBERT\footnote{https://huggingface.co/mental/mental-bert-base-uncased} (mental-roberta-base\footnote{https://huggingface.co/mental/mental-roberta-base})} - is a domain-adapted BERT model pretrained specifically on large-scale mental health-related Reddit data. Unlike general-purpose BERT and RoBERTa, mentalBERT’s pretraining corpus is drawn from online communities where users disclose experiences of depression, stress, and suicidal ideation. This makes its language representations closely aligned with the discourse and symptom expressions typical of peer-support forums. Prior work has shown that such domain-specific pretraining improves robustness and reliability, as the model learns subtle linguistic markers, colloquial expressions, and context-specific cues that general models often overlook \cite{alsentzer2019publicly}. In this study, we fine-tune the best-performing variant, mental/mental-roberta-base, to evaluate whether leveraging both large-scale pretraining and mental health–oriented domain adaptation can further enhance classification performance across depression, anxiety, PTSD, stress, suicidal ideation, and neutral content.

\textbf{RoBERTa (all-roberta-large-v1)}\footnote{https://huggingface.co/sentence-transformers/all-roberta-large-v1} - is an optimized variant of BERT that benefits from larger-scale pretraining and the removal of the next sentence prediction objective. The all-roberta-large-v1 version is further adapted in the Sentence-Transformers framework to generate high-quality sentence embeddings. It is based on the RoBERTa-large backbone with 24 layers, 1024 hidden dimensions, 16 attention heads, and approximately 355 million parameters, producing 1024-dimensional normalized embeddings. In this study, the model is fine-tuned for multi-class classification of mental health conditions, leveraging its strong representational power to capture subtle linguistic distinctions across categories. Input texts are tokenized with a maximum sequence length of 512, and training is carried out using the Hugging Face Trainer API with weighted cross-entropy loss to address class imbalance. Early stopping is applied based on macro F1 to prevent overfitting, while gradient accumulation enables effective large-batch training despite GPU memory constraints. Model performance is monitored across training, validation, and test splits, with both overall and label-wise metrics reported. The best checkpoint is retained, and evaluation includes classification reports and confusion matrices to capture detailed error patterns. In this paper, we refer to this fine-tuned model as \textbf{`multiMentalRoBERTa'}.

All models in this study were fine-tuned on a Dell Precision 3460 workstation equipped with an Intel Core i7-14700 processor (2.10 GHz, 20 cores), 32 GB RAM, and an NVIDIA RTX 2000 Ada GPU. This configuration provided sufficient computational resources to support model fine-tuning. 

\begin{table*}[tb]
\caption{Comparison of Model Performance (Macro Scores) With and Without Stress}
\label{table-model-performance-stress-vs-nostress}
\centering
\begin{tabular}{llcccc|cccc}
\toprule
\multirow{3}{*} {Group} & \multirow{3}{*} {Model} & \multicolumn{4}{c|}{6 Class (With Stress)} & \multicolumn{4}{c}{5 Class (Without Stress)} \\
\cmidrule(lr){3-6} \cmidrule(lr){7-10}
 &  & Accuracy & Precision & Recall & F1 & Accuracy & Precision & Recall & F1 \\
\midrule
\multirow{2}{*}{Traditional ML} 
 & Logistic Regression   & 0.792 & 0.787 & 0.700 & 0.729 & 0.800 & 0.801 & 0.741 & 0.763 \\
 & Linear SVM            & 0.792 & 0.795 & 0.702 & 0.735 & 0.792 & 0.779 & 0.729 & 0.748 \\
\midrule
\multirow{4}{*}{Transformer Fine-Tuned} 
 & BERT    & 0.792 & 0.740 & 0.768 & 0.751 & 0.809 & 0.805 & 0.819 & 0.808 \\
 & SBERT         & 0.792 & 0.734 & 0.773 & 0.748 & 0.783 & 0.773 & 0.789 & 0.778 \\
 & MentalBERT    & 0.862 & 0.820 & 0.837 & 0.826 & 0.870 & 0.864 & 0.860 & 0.859 \\
 & RoBERTa & 0.879 & 0.836 & \textbf{0.843} & \textbf{0.839} & 0.881 & 0.875 & \textbf{0.869} & \textbf{0.870} \\
\midrule
\multirow{6}{*}{Prompting}
\multirow{3}{*}{\hspace{1pt} (Zero-shot)}
 & DeepSeek              & 0.591 & 0.572 & 0.664 & 0.579 & 0.715 & 0.690 & 0.736 & 0.694 \\
 & GPT-4.1               & 0.575 & 0.572 & 0.656 & 0.562 & 0.700 & 0.703 & 0.731 & 0.692 \\
 & GPT-5                 & 0.571 & 0.579 & 0.661 & 0.561 & 0.664 & 0.684 & 0.720 & 0.662 \\
\cmidrule(l){2-10}
\multirow{1}{*}{\hspace{37pt}}
\multirow{3}{*}{(Few-shot)}
 & DeepSeek              & 0.594 & 0.571 & 0.663 & 0.577 & 0.704 & 0.683 & 0.741 & 0.688 \\
 & GPT-4.1               & 0.584 & 0.579 & 0.667 & 0.572 & 0.696 & 0.696 & 0.725 & 0.683 \\
 & GPT-5                 & 0.575 & 0.580 & 0.657 & 0.560 & 0.674 & 0.693 & 0.726 & 0.673 \\
\bottomrule
\end{tabular}
\end{table*}

\subsection{Prompt Engineering}\label{sec-prompt}

Large language models are evaluated through structured prompts designed to guide classification of mental health texts. Each prompt begins with a clear task definition, contextual notes to prevent medical advice or crisis instructions, and an explicit list of predefined labels. To reduce ambiguity, rules for handling uncertainty are also included, instructing the model to choose the closest category or return none when no signal of mental health content is present. In addition, the models are explicitly instructed to assign an Unknown label when the input cannot be confidently classified, a strategy shown to reduce hallucinations and improve reliability in prior work \cite{islam2025llm}. The output format is constrained to a strict schema (id - predicted\_label - text) to ensure reproducible and interpretable evaluation.

Two prompting strategies are employed. In the zero-shot setting, the model receives only the task description, label definitions, and formatting rules before classifying unseen test inputs, relying entirely on pretrained knowledge. In the few-shot setting, additional guidance is provided in the form of labeled examples. Specifically, three representative examples from each class are included within the prompt to demonstrate the labeling scheme. These few-shot prompts enable the model to capture category-specific language patterns more effectively and apply them to new instances.

All prompting experiments are executed through AI agents (GPT\footnote{https://openai.com/api/} and DeepSeek\footnote{https://api-docs.deepseek.com/}) using API endpoints. Generation parameters are fixed with a batch size of 5, temperature of 0.0, and top-p of 1.0 to encourage deterministic outputs while permitting limited diversity \cite{alammar2024hands}. An exception is GPT-5, which currently supports only a fixed temperature of 1.0 (see Code Listing \ref{lst:llm_error}), resulting in less constrained outputs. Together, these prompting strategies and parameter configurations enable systematic evaluation of model reasoning without task-specific fine-tuning.

\begin{lstlisting}[caption={Example of runtime error from the ChatGPT API}, label={lst:llm_error}]
RuntimeError: LLM API error with model 'gpt-5': 400 {
  "error": {
    "message": "Unsupported value: 'temperature' does not support 0.0 with this model. Only the default (1) value is supported.",
    "type": "invalid_request_error",
    "param": "temperature",
    "code": "unsupported_value"
  }
}
\end{lstlisting}

\section{Result Analysis}\label{result-analysis}

Model performance is assessed using four standard metrics that are widely applied in classification tasks: Accuracy, Precision, Recall, and F1-score. 
For mental health disorder detection, these metrics carry particular significance. The dataset exhibits class imbalance, with larger classes such as Stress and Depression dominating, while minority classes such as Anxiety and PTSD remain underrepresented. In such contexts, Accuracy alone can be misleading, as models may achieve high overall scores while failing to identify smaller but critical categories. To address this, macro-averaged scores are reported, giving equal weight to each class regardless of size and ensuring that minority categories are fairly evaluated. 

Among the metrics, Recall is emphasized as especially important in this domain. From a machine learning perspective, improving Recall helps ensure that instances of mental health conditions are less likely to be missed, even at the expense of occasionally lowering Precision. This is critical because under-identification of conditions such as Suicidal ideation or PTSD carries greater practical risks than occasional false positives. Therefore, while all four metrics are reported for completeness, model development and comparisons prioritize gains in Recall and macro-averaged scores as indicators of meaningful progress.

\begin{table*}[tb]
\caption{Class wise Precision, Recall, and F1 Scores With and Without Stress for Fine tuned RoBERTa and MentalBERT}
\label{table-labelwise-prf-stress-vs-nostress}
\centering
\setlength{\tabcolsep}{4pt}
\begin{tabular}{lccc|ccc|ccc|ccc}
\toprule
\multirow{4}{*}{Class} 
 & \multicolumn{6}{c|}{Fine-tuned RoBERTa / multiMentalRoBERTa} 
 & \multicolumn{6}{c}{Fine-tuned MentalBERT} \\
\cmidrule(lr){2-7}\cmidrule(lr){8-13}
 & \multicolumn{3}{c|}{6 Class (With Stress)} & \multicolumn{3}{c|}{5 Class (Without Stress)} 
 & \multicolumn{3}{c|}{6 Class (With Stress)} & \multicolumn{3}{c}{5 Class (Without Stress)} \\
\cmidrule(lr){2-4}\cmidrule(lr){5-7}\cmidrule(lr){8-10}\cmidrule(lr){11-13}
 & Precision & Recall & F1 & Precision & Recall & F1 
 & Precision & Recall & F1 & Precision & Recall & F1 \\
\midrule
Stress      
 & 0.929 & 0.916 & \textbf{0.922} & \multicolumn{3}{c|}{Not Applicable}
 & 0.916 & 0.916 & 0.916 & \multicolumn{3}{c}{Not Applicable} \\
Anxiety     
 & 0.681 & 0.762 & \textbf{0.719} & 0.796 & 0.833 & \textbf{0.814} 
 & 0.623 & \textbf{0.786} & 0.695 & 0.761 & 0.833 & 0.796 \\
Depression  
 & 0.904 & \textbf{0.892} & \textbf{0.898} & 0.920 & \textbf{0.892} & \textbf{0.906} 
 & 0.899 & 0.845 & 0.871 & 0.919 & 0.875 & 0.896 \\
PTSD        
 & 0.744 & 0.707 & 0.725 & 0.944 & \textbf{0.829} & \textbf{0.883} 
 & 0.769 & \textbf{0.732} & \textbf{0.750} & 0.943 & 0.805 & 0.868 \\
Suicidal    
 & 0.775 & \textbf{0.821} & \textbf{0.798} & 0.729 & 0.833 & 0.778 
 & 0.728 & 0.798 & 0.761 & 0.710 & \textbf{0.845} & 0.772 \\
None        
 & 0.986 & 0.958 & 0.971 & 0.986 & 0.958 & 0.971 
 & 0.985 & 0.944 & 0.964 & 0.985 & 0.944 & 0.964 \\
\midrule
Accuracy    
 & \multicolumn{3}{c|}{0.880} & \multicolumn{3}{c|}{0.881} 
 & \multicolumn{3}{c|}{0.862} & \multicolumn{3}{c}{0.870} \\
Macro avg   
 & 0.836 & \textbf{0.843} & \textbf{0.839} & 0.875 & \textbf{0.869} & \textbf{0.870} 
 & 0.820 & 0.837 & 0.826 & 0.864 & 0.860 & 0.859 \\
Weighted avg
 & 0.882 & 0.880 & 0.880 & 0.887 & 0.881 & 0.883 
 & 0.869 & 0.862 & 0.864 & 0.879 & 0.870 & 0.873 \\
\bottomrule
\end{tabular}
\end{table*}

\begin{table*}[tb]
\caption{Class-wise Confusion Matrices With and Without Stress (Fine-tuned RoBERTa / multiMentalRoBERTa)}
\label{table-confusion-stress-vs-nostress}
\centering
\scriptsize
\begin{tabular}{lcccccc|ccccc}
\toprule
\multirow{3}{*}{True Label} & \multicolumn{6}{c|}{Predicted Label (With Stress)} & \multicolumn{5}{c}{Predicted Label (Without Stress)} \\
\cmidrule(lr){2-7}\cmidrule(lr){8-12}
 & Stress & Anxiety & Depression & PTSD & Suicidal & None & Anxiety & Depression & PTSD & Suicidal & None \\
\midrule
Stress     & 208 & 7   & 3   & 7  & 1   & 1  & \multicolumn{5}{c}{Not Applicable} \\
Anxiety    & 5   & 32  & 2   & 3  & 0   & 0  & 35 & 3   & 2  & 2  & 0  \\
Depression & 2   & 4   & 207 & 0  & 19  & 0  & 3  & 207 & 0  & 22 & 0  \\
PTSD       & 7   & 4   & 1   & 29 & 0   & 0  & 4  & 2   & 34 & 0  & 1  \\
Suicidal    & 1   & 0   & 14  & 0  & 69  & 0  & 1  & 13  & 0  & 70 & 0  \\
None       & 1   & 0   & 2   & 0  & 0   & 68 & 1  & 0   & 0  & 2  & 68 \\
\bottomrule
\end{tabular}
\end{table*}

\begin{table}[tb]
\caption{Robustness Analysis of Multiclass Classifiers on  eRisk 2025 Dataset \cite{parapar2025erisk} }
\label{table-erisk-dep-robust}
\centering
\begin{tabular}{lcc|cc}
\toprule
\multirow{3}{*}{Model} 
& \multicolumn{2}{c|}{6 Class} 
& \multicolumn{2}{c}{5 Class} \\
\cmidrule(lr){2-3}\cmidrule(lr){4-5}
& Recall & F1 & Recall & F1 \\
\midrule
multiMentalRoBERTa     & 0.860 & 0.699 & 0.920 & 0.692 \\
Fine-tuned MentalBERT  & 0.860 & 0.672 & 0.900 & 0.687 \\
\bottomrule
\multicolumn{5}{l}{*Only `depression' class performance is shown}
\end{tabular}
\end{table}

The results in Table~\ref{table-model-performance-stress-vs-nostress} highlight clear differences across model groups. Traditional machine learning baselines, including Logistic Regression and Linear SVM, achieve moderate performance with F1 scores around 0.73 - 0.76. In contrast, transformer-based fine-tuned models perform substantially better. While domain-specific MentalBERT provides competitive improvements (F1 = 0.826 with stress, 0.859 without stress), multiMentalRoBERTa (fine-tuned RoBERTa) stands out as the strongest model, achieving an F1 score of 0.839 in the six-class setup and 0.870 in the five-class setup, underscoring the benefits of large-scale pretraining combined with task-specific fine-tuning. Zero-shot and few-shot prompting models perform noticeably lower, though still above chance, showing that large language models can generalize without fine-tuning but remain less reliable for sensitive classification tasks.

The effect of removing the Stress class is also evident across nearly all models. Both traditional machine learning methods and transformers show improved performance when evaluated on the five-class setting. For example, Logistic Regression improves from F1 = 0.729 to 0.763, MentalBERT improves from 0.826 to 0.859 and multiMentalRoBERTa (fine-tuned RoBERTa) rises from 0.839 to 0.870. This confirms the findings from data exploration that Stress is linguistically diffuse and overlaps with several other categories, making it harder to classify consistently. Removing this category reduces ambiguity and leads to more stable performance across the remaining conditions.

A closer look at Recall further emphasizes the advantages of fine-tuned transformers. multiMentalRoBERTa achieves the highest Recall overall, reaching 0.843 with stress included and 0.869 when stress is excluded. Fine-tuned MentalBERT also provides solid recall values (0.837 with stress and 0.862 without stress), but it remains slightly behind multiMentalRoBERTa. Other fine-tuned models such as BERT and SBERT also outperform prompting-based approaches. In comparison, zero-shot and few-shot prompting models achieve Recall values in the range of 0.65–0.74, demonstrating limited sensitivity to minority classes despite their convenience for rapid prototyping.

To further analyze the differences between transformer-based models, Table~\ref{table-labelwise-prf-stress-vs-nostress} presents class-wise Precision, Recall, and F1 scores for multiMentalRoBERTa (fine-tuned RoBERTa) and fine-tuned MentalBERT. Across nearly all classes, multiMentalRoBERTa achieves higher F1 scores than MentalBERT, particularly for Anxiety, Depression, and PTSD, where the improvements are most pronounced. For example, in the five-class setup, multiMentalRoBERTa attains an F1 of 0.814 for Anxiety and 0.906 for Depression, compared to 0.796 and 0.896 respectively for MentalBERT. While MentalBERT shows competitive recall for certain categories such as Suicidal ideation, the overall pattern demonstrates that multiMentalRoBERTa provides more consistent and stronger performance across categories. This reinforces the conclusion that multiMentalRoBERTa not only excels in macro-averaged metrics but also dominates in class-wise evaluations, underscoring its robustness across both majority and minority mental health conditions.

\begin{table*}[tb]
\centering
\caption{Top 5 meaningful words driving suicide classification across error types}
\label{table:suicide-explain-words}
\begin{tabular}{lcc}
\toprule
\textbf{Category} & \textbf{Toward Suicide (Positive Drivers)} & \textbf{Away from Suicide (Negative Drivers)} \\
\midrule
False Negatives (Suicide $\rightarrow$ Other) &
cutting, insane, dead, therapist, thought &
lived, broke, stuttering, school, better \\
\midrule
False Positives (Depression $\rightarrow$ Suicide) &
suicidal, suicide, killing, hotline, feelings &
manipulative, embarrassing, care, excited, option \\
\midrule
True Positives (Suicide $\rightarrow$ Suicide) &
alone Dying, suicide, kill, death, fucking &
goodbye, family, problems, assaulted, account \\
\bottomrule
\end{tabular}
\end{table*}

Since no benchmark dataset exists for evaluating robustness in our five-class and six-class multiclass settings, we conducted additional experiments by collapsing the problem into a binary classification of depression versus non-depression. While benchmark datasets such as eRisk exist for binary depression detection, they do not extend to multiclass scenarios. To fill this gap, we used the eRisk 2025 dataset (Task 2)\cite{parapar2025erisk} to test robustness, aggregating posts into binary labels. Because our models were originally fine-tuned for multiclass classification rather than binary tasks, performance was lower in this setting. Nonetheless, as shown in Table~\ref{table-erisk-dep-robust}, our multiMentalRoBERTa model consistently achieved stronger recall and F1 than fine-tuned MentalBERT, demonstrating superior robustness.

Overall, four conclusions emerge. First, multiMentalRoBERTa establishes itself as the best-performing model, balancing Precision, Recall, and F1 effectively. Second, excluding Stress improves classification consistency, reflecting its overlapping linguistic nature. Third, traditional machine learning models remain useful as baselines but are clearly surpassed by modern transformers. Finally, zero-shot and few-shot prompting show promise but require careful refinement before they can be relied upon in safety-critical mental health applications. Therefore, all subsequent analyses in this study are conducted using the outputs of the multiMentalRoBERTa model.

The class-wise results in Table~\ref{table-labelwise-prf-stress-vs-nostress} and Table~\ref{table-confusion-stress-vs-nostress} provide deeper insight into model behavior across individual categories. When Stress is included as a separate label, it achieves strong Precision (0.929) and Recall (0.916), indicating that the majority of stress-related posts are captured accurately. However, the confusion matrix shows that several Anxiety and PTSD cases are misclassified as Stress, reflecting the broad and overlapping nature of stress expressions. After the Stress class is removed, these ambiguous cases are redistributed across the remaining categories, leading to substantial improvements. For instance, Anxiety F1 rises from 0.719 to 0.814, and PTSD F1 increases from 0.725 to 0.883. This demonstrates that Stress serves as a confounding label that pulls examples away from more specific conditions, thereby reducing classification clarity when included.

The interplay between Depression and Suicidal ideation is particularly noteworthy. In the six-class setting, 19 depression cases are misclassified as suicidal and 14 suicidal cases are classified as depression. These bidirectional errors align with the strong semantic connection between depressive symptoms and suicidal ideation. Despite this overlap, Depression maintains high performance across both setups, with F1 hovering around 0.898 - 0.906. Suicidal shows robust Recall (0.821 with stress, 0.833 without stress), which is encouraging since recall is the priority metric for safety-critical conditions. However, its Precision decreases from 0.775 to 0.729 after stress removal, indicating an increase in false positives once the stress category is no longer available to absorb borderline cases. This trade-off highlights the challenges of distinguishing suicidal content from severe depressive language.

The None class is consistently well identified, with Precision at 0.986 and Recall at 0.958 across both conditions. Its linguistic distinctiveness from mental health-related text makes it relatively easy to classify correctly. Overall, the per-class results confirm three patterns: (1) Stress overlaps with multiple categories and weakens the separability of specific conditions, (2) Depression and Suicidal are strongly interlinked, and (3) Anxiety and PTSD benefit considerably when stress is excluded as a competing class.

These per-class results are consistent with the earlier findings from the Data Exploration stage. In the cluster distribution (Table~\ref{table-cluster-distribution}) and correlation matrix (Table~\ref{table-correlation-matrix}), Depression and Suicidal were shown to be highly correlated (0.929), which directly corresponds to the misclassification patterns observed in the confusion matrix. Similarly, Anxiety and PTSD exhibited a strong correlation (0.843) during embedding analysis, and their improved scores after stress removal demonstrate how eliminating the confounding effect of stress clarifies their boundaries. The widespread distribution of Stress across clusters in Figure~\ref{Fig:class-cluster} also aligns with its role as a diffuse category that overlaps with several conditions, explaining why its removal improves overall macro scores.

\begin{figure}[tb]
\centerline{\includegraphics[width=0.45\textwidth]{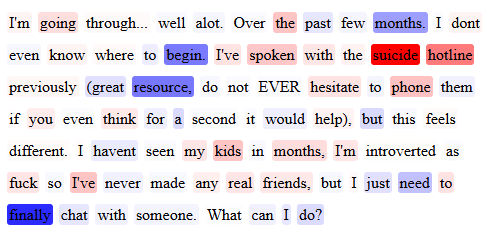}}
\caption{Sample Example of Depression predicted as Suicidal [Red = pushes toward the predicted class (supportive evidence), Blue = pushes against the predicted class (counter-evidence), Light vs. Dark = strength of the effect.]}
\label{Fig:sample-example}
\end{figure}

In summary, the quantitative evidence from the confusion matrices and class-wise metrics reinforces the qualitative insights gained from embedding-based exploration. The consistency between these stages strengthens confidence in the interpretation that Stress introduces noise, Depression and Suicidal are semantically close, and Anxiety and PTSD share linguistic features that complicate classification.

\begin{table*}[ht]
\centering
\caption{Representative KeyBERT phrases by bucket (top 10 per category after removing generic/stop phrases).}
\label{tab:keybert-meaningful-phrases}
\begin{tabular}{p{3cm}p{14cm}}
\toprule
\textbf{Category} & \textbf{Top phrases} \\
\midrule
Suicide $\rightarrow$ Depression (False Negatives) &
cutting felt good, days free cutting, suicide ideation, suicide attempts, feelings thoughts suicide, depressed survived suicide, want life anymore, feel like drowning, dead want drown, feeling hopeless \\

\midrule
Depression $\rightarrow$ Suicide (False Positives) &
suicide hotline, emotions directing death, dying want thoughts, failure deserve die, easier just end, extremely depressive episode, just want happy, feel like mistake, literally don care, help suicidal friends \\

\midrule
Depression $\rightarrow$ Depression (True Negatives) &
procrastination depression help, depression make forgetful, post graduation depression, helping depression, oils depression anxiety, prescribed severe depression, dealing major depressive, functional depression, stop negative thoughts, feel better \\

\midrule
Suicide $\rightarrow$ Suicide \hspace{5px} (True Positives) &
suicide selected hypothermia, thinking time death, jump cliff, kill 20th birthday, suicidal thoughts, I've attempted suicide, I've contemplated suicide, plan jumping car, die old room, like screaming help \\
\bottomrule
\end{tabular}
\end{table*}

\section{Explainability  Analysis}\label{explainability-analysis}
The explainability analysis primarily concentrates on the Suicidal ideation class. This category is of particular importance since identifying early signs of suicidal tendency is critical for timely intervention and has greater practical implications than other conditions. While overall model performance improves after excluding the Stress class, suicidal ideation remains a class that requires closer examination to ensure reliable detection. Focusing on this class in explainability analysis allows us to better understand which linguistic patterns the model leverages when making predictions and how these patterns align with clinically meaningful cues.

A key challenge for this category lies in the data itself. As described in the Data Collection section, the annotations are based on subreddit membership and user self-disclosure, which introduces inherent noise into the labels. Such reliance on self-reporting can lead to inconsistencies that affect both training and evaluation. Therefore, explainability methods are applied not only to interpret model behavior but also to assess whether the model captures genuine suicidal indicators or patterns influenced by noisy labeling.

For the explainability analysis, two complementary methods are applied to better understand how the model interprets suicidal ideation. The first uses Layer Integrated Gradients (LIG)\footnote{https://captum.ai/api/layer.html\#layer-integrated-gradients}
 with the multiMentalRoBERTa classifier to attribute importance scores to tokens, showing which words push predictions toward or away from Suicidal. Representative cases include true positives (correctly identified suicidal posts), false negatives (missed suicidal posts), and false positives (depressive posts misclassified as suicidal). LIG provides word-level visualizations and aggregated rankings of influential terms, while overlap analysis between true positives and false positives highlights sources of confusion between depression and suicidal ideation. This reveals whether the model’s reasoning aligns with meaningful suicidal indicators. Figure~\ref{Fig:sample-example} illustrates an example where a post originally labeled as Depression is predicted as Suicidal. While technically a misclassification, this outcome is clinically favorable since the post contains explicit references to a suicide hotline and prolonged emotional distress. Labeling such sensitive content as suicidal ensures that high-risk cases are not overlooked, highlighting the model’s utility in prioritizing safety-critical intervention. The most influential words identified through Layer Integrated Gradients are shown in Table~\ref{table:suicide-explain-words}, with the top five drivers reported per outcome category. For false negatives, terms like cutting, dead, and therapist push the model toward suicide but are diluted by context. False positives are often triggered by explicit words such as suicidal and killing, while terms like manipulative or embarrassing counteract them. True positives are dominated by cues like suicide, kill, and death, whereas words such as family and goodbye act as negative drivers. These patterns reveal how lexical cues shape predictions and explain errors in borderline cases between depression and suicidal ideation.

The second method uses KeyBERT\footnote{https://pypi.org/project/keybert/}
 to extract key phrases, shifting focus from model reasoning to the text’s linguistic features. Across true positives, false negatives, true negatives, and false positives, salient phrases are aggregated by frequency and relevance, revealing expressions that characterize correctly detected suicidal posts, those often missed, and depressive phrases that drive misclassification. Representative phrases extracted by KeyBERT are shown in Table~\ref{tab:keybert-meaningful-phrases}, with the top ten reported for each outcome bucket. In false negatives (Suicide $\rightarrow$ Depression), explicit phrases like cutting felt good, suicide ideation, and feeling hopeless show how strong suicidal cues can still be overlooked due to context or label noise. False positives (Depression $\rightarrow$ Suicide) include terms such as suicide hotline or help suicidal friends, where discussions of suicide risk trigger misclassification despite non-suicidal intent. True negatives feature general depressive language like functional depression and feel better, while true positives capture clear suicidality with phrases like suicidal thoughts and plan jumping car. Together with token-level attributions from LIG, these phrase-level patterns provide a richer view of how suicidal ideation is expressed and why depression–suicide boundaries remain difficult..

\section{Bias, Fairness \& Safety } \label{sec-bias}
Mental health AI classification systems face technical and ethical challenges related to bias. As the American Psychological Association \cite{American-Psychological-Association} noted, AI must be developed and deployed with attention to transparency, informed consent, and the mitigation of systematic biases that could worsen disparities in psychological care. Training data drawn from narrow platforms or self-reported labels may embed cultural and linguistic biases that reduce generalizability across diverse populations. To promote fairness, we collected data from multiple sources, including the Stress Annotated Dataset (SAD), Reddit corpora, and neutral forums (see Section \ref{sec-data-collection}), ensuring broader representation and reducing overfitting to a single community. We also addressed ethical concerns (see section \ref{sec-data-collection}) by applying de-identification and safeguards against re-identification, aligning the model with responsible mental health research practices.

Finally, given the high stakes of mental health applications, safety remains central. False negatives in suicidal ideation or acute distress carry severe consequences; therefore, our models prioritize recall. Human-in-the-loop protocols ensure that AI complements, rather than replaces, professional judgment, reinforcing trust and accountability in clinical and peer-support contexts.

\section{Conclusion}
This study introduced multiMentalRoBERTa, a fine-tuned RoBERTa model for multiclass classification of mental health disorders. By integrating diverse datasets and exploring overlaps across conditions, the model demonstrated superior performance over baselines and domain-specific transformers, achieving robust results particularly when stress was excluded as a confounding class. Explainability analyses further revealed meaningful cues for depression and suicidal ideation, underscoring the model’s reliability and safety in sensitive contexts. Overall, multiMentalRoBERTa provides a lightweight, interpretable, and deployable solution for enhancing early detection and support in mental health platforms.

Looking ahead, future work will focus on incorporating human evaluation to assess the interpretability and clinical relevance of model outputs, ensuring that explanations align with the judgments of mental health professionals and peer-support specialists. Additionally, exploring clinical usage scenarios, such as integration into triage systems, decision-support tools, will be critical for translating research findings into real-world applications.

\section*{Acknowledgment}
We gratefully acknowledge the support of the Northwestern Mutual Data Science Institute (NMDSI) for funding this project.

\bibliographystyle{IEEEtran}
\bibliography{Ref}
\vspace{12pt}

\end{document}